\documentclass[11pt,a4paper]{article}
\usepackage[hyperref]{naaclhlt2019}
\usepackage{times}
\usepackage{latexsym}
\usepackage{amsmath,amssymb}
\usepackage{tabularx}
\usepackage{graphicx}
\usepackage{subcaption}
\usepackage{adjustbox}
\usepackage{bm}
\usepackage{multirow}

\usepackage{url}

\aclfinalcopy
\def\aclpaperid{1450}

\newcommand\blfootnote[1]{%
  \begingroup
  \renewcommand\thefootnote{}\footnote{#1}%
  \addtocounter{footnote}{-1}%
  \endgroup
}

\newcommand\possesivecite[1]{\citeauthor{#1}'s (\citeyear{#1})}
\newcommand{\wtv}{word2vec \hspace{0ex}}
\newcommand{\Sref}[1]{\S\ref{#1}}

\title{Black is to Criminal as Caucasian is to Police: \\ Detecting and Removing Multiclass Bias in Word Embeddings}

\author{Thomas Manzini$^{\dagger}$* , Yao Chong Lim$^{\ddagger}$* , Yulia Tsvetkov$^{\ddagger}$, Alan W Black$^{\ddagger}$\\
  Microsoft AI Development Acceleration Program$^{\dagger}$, Carnegie Mellon University$^{\ddagger}$ \\
  {\tt Thomas.Manzini@microsoft.com, \{yaochonl,ytsvetko,awb\}@cs.cmu.edu}}
  
\date{}

\begin{document}
\maketitle
\begin{abstract}
\blfootnote{* Equal contributions\newline\indent\hspace{2.6mm}${\dagger}$ Work done while at CMU and The Microsoft AI Development Acceleration Program}
Online texts---across genres, registers, domains, and styles---are riddled with human stereotypes, expressed in overt or subtle ways. 
Word embeddings, trained on these texts, perpetuate and amplify these stereotypes, and propagate biases to machine learning models that use word embeddings as features.  
In this work, we propose a method to debias word embeddings in multiclass settings such as race and religion, extending the work of \citep{bolukbasi2016man} from the binary setting, such as binary gender.
Next, we propose a novel methodology for the evaluation of multiclass debiasing. 
We demonstrate that our multiclass debiasing is robust and maintains the efficacy in standard NLP tasks.  
\end{abstract}

\section{Introduction}
In addition to possessing informative features useful for a variety of NLP tasks, word embeddings reflect and propagate social biases present in training corpora \citep{caliskan2017semantics,GargE3635}. Machine learning systems that use embeddings can further amplify biases \citep{barocas2016big,zhao2017men}, discriminating against users, particularly those from disadvantaged social groups. 

\citep{bolukbasi2016man} introduced a method to \emph{debias} embeddings by removing components that lie in stereotype-related embedding subspaces. 
They demonstrate the effectiveness of the approach by removing gender bias from \wtv embeddings \citep{mikolov2013efficient}, preserving the utility of embeddings and potentially alleviating biases in downstream tasks. However, this method was only for \emph{binary} labels (e.g., male/female), whereas most real-world demographic attributes, including gender, race, religion, are not binary but continuous or categorical, with more than two categories.


\begin{table}
\centering
\begin{tabular}{|ll|}
\hline
\multicolumn{2}{|c|}{\textbf{Racial Analogies}}        \\ \hline
black $\rightarrow$ homeless & caucasian $\rightarrow$ servicemen   \\
caucasian $\rightarrow$ hillbilly & asian $\rightarrow$ suburban \\
asian $\rightarrow$ laborer & black $\rightarrow$ landowner        \\ \hline

\multicolumn{2}{|c|}{\textbf{Religious Analogies}}          \\ \hline            
jew $\rightarrow$ greedy & muslim $\rightarrow$ powerless \\
christian $\rightarrow$ familial & muslim $\rightarrow$ warzone   \\
muslim $\rightarrow$ uneducated & christian $\rightarrow$ intellectually \\ \hline
\end{tabular}
\caption{Examples of racial and religious biases in analogies generated from word embeddings trained on the Reddit data from users from the USA.}
\label{table:biasanalogy}
\end{table}


In this work, we show a generalization of \possesivecite{bolukbasi2016man} which enables \emph{multiclass} debiasing, while preserving utility of embeddings (\Sref{sec:multiclass-debiasing}). 
We train \wtv embeddings using the Reddit L2 corpus \citep{rabinovich2018} and apply multiclass debiasing using lexicons from studies on bias in NLP and social science (\Sref{sec:lexicons}). 
We introduce a novel metric for evaluation of bias in collections of word embeddings (\Sref{sec:eval}).
Finally, we validate that the utility of debiased embeddings in the tasks of part-of-speech (POS) tagging, named entity recognition (NER), and POS chunking is on par with off-the-shelf embeddings. 



\section{Background}
As defined by \citep{bolukbasi2016man}, debiasing word embeddings in a binary setting requires identifying the bias subspace of the embeddings. Components lying in that subspace are then removed from each embedding.

\subsection{Identifying the bias subspace}\label{subsection:id_bias}
\citep{bolukbasi2016man} define the gender subspace using \emph{defining sets} of words, where the words in each set represent different ends of the bias. For example, in the case of gender, one defining set might be the gendered pronouns \{\textit{he}, \textit{she}\} and another set might be the gendered nouns \{\textit{man}, \textit{woman}\}.
The gender subspace is then computed from these defining sets by 1) computing the vector differences of the word embeddings of words in each set from the set's mean, and 2) taking the most significant components of these vectors.


\subsection{Removing bias components}
\label{section:debiasing_bolukbasi}
Following the identification of the gender subspace, one can apply hard or soft debiasing \citep{bolukbasi2016man} to completely or partially remove the subspace components from the embeddings.

\subsubsection*{Hard debiasing}


Hard debiasing (also called ``Neutralize and Equalize'') involves two steps. First, bias components are removed from words that are not gendered and should not contain gender bias (e.g., \textit{doctor}, \textit{nurse}), and second, gendered word embeddings are centered and their bias components are equalized. For example, in the binary case, \textit{man} and \textit{woman} should have bias components in opposite directions, but of the same magnitude. Intuitively, this then ensures that any neutral words are equidistant to any biased words with respect to the bias subspace.

More formally, to \emph{neutralize}, given a bias subspace $\mathcal{B}$ spanned by the vectors $\{\bm{b_1}, \bm{b_2}, ..., \bm{b_k}\}$, we compute the component of each embedding in this subspace:
\begin{align}\label{eqn:bias_component}
    \mathbf{w}_{\mathcal{B}} = \sum_{i=1}^k \langle \mathbf{w}, \bm{b_i} \rangle \bm{b_i}
\end{align}
We then remove this component from words that should be bias-neutral and normalize to get the debiased embedding:
\begin{align}\label{eqn:debias}
    \mathbf{w'} = \frac{\mathbf{w} - \mathbf{w}_{\mathcal{B}}}{\lVert \mathbf{w} - \mathbf{w}_{\mathcal{B}} \rVert}
\end{align}
To \emph{equalize} the embeddings of words in an equality set $E$, let $\bm{\mu} = \frac{1}{|E|} \sum_{\mathbf{w} \in E} \mathbf{w}$ be the mean embedding of the words in the set and $\bm{\mu}_{\mathcal{B}}$ be its component in the bias subspace as calculated in Equation~\ref{eqn:bias_component}. Then, for $\mathbf{w} \in E$,
\begin{align}\label{eqn:equalize}
    \mathbf{w'} = (\bm{\mu} - \bm{\mu}_{\mathcal{B}}) + \sqrt{1 - \lVert \bm{\mu} - \bm{\mu}_{\mathcal{B}} \rVert^2} \frac{\mathbf{w}_{\mathcal{B}} - \bm{\mu}_{\mathcal{B}}}{\lVert \mathbf{w}_{\mathcal{B}} - \bm{\mu}_{\mathcal{B}} \rVert}
\end{align}
Note that in both Equations~\ref{eqn:debias} and \ref{eqn:equalize}, the new embedding has unit length.

\subsubsection*{Soft debiasing}

Soft debiasing involves learning a projection of the embedding matrix that preserves the inner product between biased and debiased embeddings while minimizing the projection onto the bias subspace of embeddings that should be neutral.

Given embeddings $\mathbf{W}$ and $\mathbf{N}$ which are embeddings for the whole vocabulary and the subset of bias-neutral words respectively, and the bias subspace $\mathcal{B}$ obtained in Section~\ref{subsection:id_bias}, soft debiasing seeks for a linear transformation $A$ that minimizes the following objective:
\begin{align}\label{eqn:soft_debias}
    \begin{split}
    \lVert (\mathbf{AW})^{\intercal}(\mathbf{AW}) - \mathbf{W}^{\intercal}\mathbf{W} \rVert^2_F \\
    +\, \lambda \lVert (\mathbf{AN})^{\intercal}(\mathbf{A}\mathcal{B}) \rVert^2_F
    \end{split}
\end{align}
Minimizing the first term preserves the inner product after the linear transformation $\mathbf{A}$, and minimizing the second term minimizes the projection onto the bias subspace $\mathcal{B}$ of embeddings. $\lambda \in \mathbb{R}$ is a tunable parameter that balances the two objectives.

\section{Methodology}
\label{sec:multiclass-debiasing}
We now discuss our proposed extension of word embedding debiasing to the multiclass setting.

\subsection{Debiasing}

\label{section:debiasing}

As in the binary setting, debiasing consists of two steps: identifying the ``bias subspace'' and removing this component from the set of embeddings.


\subsubsection*{Identifying the bias subspace}

The core contribution of our work is in \emph{identifying} the ``bias subspace'' in a multiclass setting; if we can identify the bias subspace then prior work can be used for multiclass debiasing.

Past work has shown that it is possible to linearly separate multiple social classes based on components of word embeddings \citep{GargE3635}. Based on this we hypothesize that there exists some component of these embeddings which can capture multiclass bias. While a multiclass problem is inherently not a linearly separable problem, a one versus rest classifier is. Following from this, the computation of a multiclass bias subspace does not have any linear constraints, though it does come with a loss of resolution.
As a result we can compute the principal components required to compute the ``bias subspace'' by simply adding an additional term for each additional bias class to each defining set.

Formally, given defining sets of word embeddings $D_1, D_2, ..., D_n$, let the mean of the defining set $i$ be $\bm{\mu}_i = \frac{1}{|D_i|} \sum_{\mathbf{w} \in D_i} \mathbf{w}$, where $\mathbf{w}$ is the word embedding of $w$. Then the bias subspace $\mathcal{B}$ is given by the first $k$ components of the following principal component analysis (PCA) evaluation: 
\begin{align}
\mathbf{PCA} \left( \bigcup_{i=1}^n \bigcup_{\mathbf{w} \in D_i} \mathbf{w} - \bm{\mu}_i \right)
\end{align}
The number of components $k$ can be empirically determined by inspecting the eigenvalues of the PCA, or using a threshold. Also, note that the defining sets do not have to be the same size. We discuss the robustness of this method later. 

\begin{table}
\centering
\begin{tabular}{|l|l|l|}
\hline
\textbf{Gender Debiasing} & MAC & $p$-Value \\ \hline    
Biased & 0.623 & N/A \\
Hard Debiased & 0.700 & 3.2e-10\\
Soft Debiased ($\lambda$ = 0.2) & 0.747 & 1.711e-12\\ \hline

\textbf{Race Debiasing} & MAC & $p$-Value \\ \hline
Biased & 0.892 & N/A \\
Hard Debiased & 0.925 & 0.0298 \\
Soft Debiased ($\lambda$ = 0.2) & 0.985 & 6.217e-05 \\ \hline

\textbf{Religion Debiasing} & MAC & $p$-Value \\ \hline           
Biased & 0.859 & N/A \\
Hard Debiased & 0.934 & 1.469e-06 \\
Soft Debiased ($\lambda$ = 0.2) & 0.894 & 0.007 \\ \hline

\end{tabular}
\caption{The associated mean average cosine similarity (MAC) (defined in Section \ref{subsection:mac}) and $p$-Values for debiasing methods for gender, race, and religious bias.}
\label{table:biasmac}
\end{table}

\subsubsection*{Removing Bias Components}
Following the identification of the bias subspace, we apply the hard Neutralize and Equalize debiasing and soft debiasing method presented in \citep{bolukbasi2016man} and discussed in Section~\ref{section:debiasing_bolukbasi} to completely or partially remove the subspace components from the embeddings.

For equalization, we take the defining sets to be the equality sets as well.

\subsection{Quantifying Bias Removal}\label{subsection:mac}
We propose a new metric for the evaluation of bias in collections of words which is simply the mean average cosine similarity (MAC). This approach is motivated by the WEAT evaluation method proposed by \citep{caliskan2017semantics} but modified for a multiclass setting. To compute this metric the following data is required: a set of target word embeddings $T$ containing terms that inherently contain some form of social bias (e.g. \{\textit{church}, \textit{synagogue}, \textit{mosque}\}), and a set $A$ which contains sets of attributes $A_1, A_2, ..., A_N$ containing word embeddings that should not be associated with any word embeddings contained in the set  $T$ (e.g. \{\textit{violent}, \textit{liberal}, \textit{conservative}\}). 

We define a function $S$ that computes the mean cosine distance between a particular target $T_i$ and all terms in a particular attribute set $A_j$:
\begin{align}
   S(\mathbf{t}, A_j) = \frac{1}{N} \sum_{\mathbf{a} \in A_j} cos(\mathbf{t}, \mathbf{a}) , 
\end{align}
where the cosine distance is:
\begin{align}
    cos(\mathbf{u}, \mathbf{v}) = 1- \frac{\mathbf{u} \cdot \mathbf{v}}{\| \mathbf{u} \|_2 \cdot \| \mathbf{v} \|_2} .
\end{align}
Finally, we define MAC as:
\begin{align}
    \textrm{MAC}(T, A) = \frac{1}{|T| |A|} \sum_{T_i \in T} \sum_{A_j \in A} S(T_i, A_j)
\end{align}
We also perform a paired $t$-test on the distribution of average cosines used to calculate the MAC. Thus we can quantify the effect of debiasing on word embeddings in $T$ and sets in $A$.



\subsection{Measuring Downstream Utility}
To measure the utility of the debiased word embeddings, we use the tasks of NER, POS tagging, and POS chunking. This is to ensure that the debiasing procedure has not destroyed the utility of the word embeddings. We evaluate test sentences that contain at least one word affected by debiasing. Additionally, we measure the change in performance after replacing the biased embedding matrix by a debiased one, and retraining the model on debiased embeddings. 

\begin{table*}[]
\centering
\begin{adjustbox}{width=\textwidth}
\begin{tabular}{|l|l|l|l|l|l|l|l|l|l|}
\hline
\multicolumn{10}{|c|}{Embedding Matrix Replacement}                                                                                                                                                                             \\ \hline
            & \multicolumn{3}{l|}{Hard Gender Debiasing}                          & \multicolumn{3}{l|}{Hard Racial Debiasing}                           & \multicolumn{3}{l|}{Hard Religious Debiasing}                        \\ \cline{2-10} 
            & NER Tagging          & POS Tagging           & POS Chunking         & NER Tagging          & POS Tagging            & POS Chunking         & NER Tagging           & POS Tagging           & POS Chunking         \\ \hline
Biased F1   & 0.9954 & 0.9657  & 0.9958 & 0.9948 & 0.9668   & 0.9958 & 0.9971  & 0.9665  & 0.9968 \\ \cline{1-1}
$\Delta$ F1        & +0.0045 & -0.0098 & +0.0041 & +0.0051 & -0.0117  & +0.0041 & +0.0103    & -0.0345 & +0.0120    \\ \cline{1-1}
$\Delta$ Precision & 0.0    & -0.0177 & 0.0    & 0.0    & -0.0208  & 0.0    & 0.0  & -0.0337 & 0.0 \\ \cline{1-1}
$\Delta$ Recall    & +0.0165 & -0.0208 & +0.0156 & +0.0186 & -0.0250  & +0.0155 & +0.00286 & -0.0174 & +0.0031 \\ \hline
            & \multicolumn{3}{l|}{Soft Gender Debiasing}                          & \multicolumn{3}{l|}{Soft Racial Debiasing}                           & \multicolumn{3}{l|}{Soft Religious Debiasing}                        \\ \cline{2-10} 
            & NER Tagging          & POS Tagging           & POS Chunking         & NER Tagging          & POS Tagging            & POS Chunking         & NER Tagging           & POS Tagging           & POS Chunking         \\ \hline
Biased F1   & 0.9952 & 0.9614  & 0.9950 & 0.9946 & 0.9612   & 0.9946 & 0.9964  & 0.9616  & 0.9961 \\ \cline{1-1}
$\Delta$ F1        & +0.0047 & -0.0102 & +0.0049 & +0.0053 & -0.0107  & +0.0053 & +0.0128     & -0.0242 & +0.0148    \\ \cline{1-1}
$\Delta$ Precision & 0.0    & -0.0202 & 0.0    & 0.0    & -0.0223  & 0.0    & 0.0  & -0.0199 & 0.0 \\ \cline{1-1}
$\Delta$ Recall    & +0.0169 & -0.0198 & +0.0187 & +0.0193 & -0.0197  & +0.0202 & +0.0035  & -0.0112 & +0.0038 \\ \hline
\multicolumn{10}{|c|}{Model Retraining} \\ \hline
            & \multicolumn{3}{l|}{Hard Gender Debiasing}                          & \multicolumn{3}{l|}{Hard Racial Debiasing}                           & \multicolumn{3}{l|}{Hard Religious Debiasing}                        \\ \cline{2-10} 
            & NER Tagging          & POS Tagging           & POS Chunking         & NER Tagging          & POS Tagging            & POS Chunking         & NER Tagging           & POS Tagging           & POS Chunking         \\ \hline
Biased F1   & 0.9954 & 0.9657  & 0.9958 & 0.9948 & 0.9668   & 0.9958 & 0.9971  & 0.9665  & 0.9968 \\ \cline{1-1}
$\Delta$ F1        & +0.0045 & -0.0137 & +0.0041 & +0.0051 & -0.0165  & +0.0041 & +0.0103      & -0.0344 & +0.0120    \\ \cline{1-1}
$\Delta$ Precision & 0.0    & -0.0259 & 0.0    & 0.0    & -0.0339  & 0.0    & 0.0  & -0.0287 & 0.0 \\ \cline{1-1}
$\Delta$ Recall    & +0.0165 & -0.0278 & +0.0156 & +0.0186 & -0.0306  & +0.0156 & +0.00286 & -0.0161 & +0.0031 \\ \hline
            & \multicolumn{3}{l|}{Soft Gender Debiasing}                          & \multicolumn{3}{l|}{Soft Racial Debiasing}                           & \multicolumn{3}{l|}{Soft Religious Debiasing}                        \\ \cline{2-10} 
            & NER Tagging          & POS Tagging           & POS Chunking         & NER Tagging          & POS Tagging            & POS Chunking         & NER Tagging           & POS Tagging           & POS Chunking         \\ \hline
Biased F1   & 0.9952 & 0.9614  & 0.9950 & 0.9946 & 0.9612   & 0.9946 & 0.9964  & 0.9616  & 0.9961 \\ \cline{1-1}
$\Delta$ F1        & +0.0047 & +0.00178 & +0.0049 & +0.0053 & -0.00119 & +0.0053 & +0.0128 & -0.0098 & +0.0148    \\ \cline{1-1}
$\Delta$ Precision & 0.0    & +0.0048  & 0.0    & 0.0    & -0.00187 & 0.0    & 0.0  & -0.0125 & 0.0 \\ \cline{1-1}
$\Delta$ Recall    & +0.0169 & +0.00206 & +0.0187 & +0.0193 & -0.00264 & +0.0202 & +0.0035  & -0.0057 & +0.0038 \\ \hline
\end{tabular}
\end{adjustbox}
\caption{The performance of embeddings the downstream tasks of NER, POS Tagging, and POS Chunking.}
\label{table:biasablation}
\end{table*}

\section{Data}

In this section we discuss the different data sources we used for our initial word embeddings, the social bias data used for evaluating bias, and the linguistic data used for evaluating the debiasing process. 

\subsection{Embedding Language Corpus}
We used the L2-Reddit corpus \citep{rabinovich2018}, a collection of Reddit posts and comments by both native and non-native English speakers. 
The native countries of post authors are determined based on their posts in country- and region-specific subreddits (such as r/Europe and r/UnitedKingdom), and other metadata such as user flairs, which serve as self-identification of the user's country of origin.

In this work, we exclusively explore data collected from the United States. This was done to leverage extensive studies of social bias done in the United States. To obtain the initial biased word embeddings, we trained \wtv  embeddings \citep{mikolov2013efficient} 
using approximately 56 million sentences. 

\subsection{Social Bias Data}
\label{sec:lexicons}
We used the following vocabularies and studies to compile lexicons for bias detection and removal.\footnote{The source and lexicons can be found here: \url{https://github.com/TManzini/DebiasMulticlassWordEmbedding/}.}

For gender, we used vocabularies created by \citep{bolukbasi2016man} and \citep{caliskan2017semantics}. 

For race we consulted a number of different sources for each race: Caucasians \citep{chung2005we, goad1998redneck}; African Americans \citep{punyanunt2008perceived, brown2005priming, chung2005we, hakanen1995emotional, welch2007black, kawai2005stereotyping}; and Asian Americans \citep{leong1990occupational, lin2005stereotype, chung2005we, osajima2005asian, GargE3635}.

Finally, for religion we used the following sources and labels:  Christians \citep{rios2015negative, zuckerman2009atheism, unnever2005turning}; Jews  \citep{dundes1971study, fetzer2000public}; and Muslims \citep{shryock2010islamophobia, alsultany2012arabs, shaheen1997arab}. 


\subsection{Downstream Tasks}
We evaluate biased and debiased word embeddings on several downstream tasks. Specifically, the CoNLL 2003 shared task \citep{tjong2003introduction} which provides evaluation data for NER, POS tagging, and POS chunking.

\section{Results and Discussion}
\label{sec:eval}
In this section we review the results of our experiments and discuss what those results mean in the context of this work. 

\subsection{Observations of Bias}
We use the analogy task from \citep{bolukbasi2016man} to demonstrate that bias exists in these word embeddings. In order to construct our analogies we trained five word2vec \citep{mikolov2013efficient} embedding spaces on the same data. We then constructed a set of analogies for each embedding space taking the intersection of these to form a working set of analogies. We performed this extra step in order to ensure the analogies were robust to perturbations in the embedding space. Following this analysis we observe that bias is present in generated analogies by viewing them directly. A small subset of these analogies are in Table \ref{table:biasanalogy} to highlight our findings.  

\subsection{Removal of Bias}
We perform our debiasing in the same manner as described in Section \ref{section:debiasing} and calculate the MAC scores and $p$-values to measure the effects of debiasing. Results are presented in Table \ref{table:biasmac}.

\textit{Does multiclass debiasing decrease bias?} We see that this debiasing procedure categorically moves MAC scores closer to 1.0. This indicates an increase in cosine distance. Further, the associated P-values indicate these changes are statistically significant. This demonstrates that our approach for multiclass debiasing decreases bias.

\subsection{Downstream Effects of Bias Removal}
The effects of debiasing on downstream tasks are shown in Table \ref{table:biasablation}. Debiasing can either help or harm performance. For POS tagging there is almost always a decrease in performance. However, for NER and POS chunking, there is a consistent increase. We conclude that these models have learned to depend on some bias subspaces differently. Note that many performance changes are of questionable statistical significance.

\textit{Does multiclass debiasing preserve semantic utility?}
We argue the minor changes in Table \ref{table:biasablation} support the preservation of semantic utility in the multiclass setting, especially compared to gender debiasing which is known to preserve utility \citep{bolukbasi2016man}.

\textit{Is the calculated bias subspace robust?}
The bias subspace is at least robust enough to support the above debiasing operations. This is shown by statistically significant changes in MAC scores.





\section{Limitations \& Future Work}
Calculating multiclass bias subspace using our proposed approach has drawbacks. For example, in the binary gender case, the extremes of bias subspace reflect extreme male and female terms. However, this is not possible when projecting multiple classes into a linear space. Thus, while we can calculate the magnitude of the bias components, we cannot measure extremes of each class.

Additionally, the methods presented here rely on words that represent biases (defining sets) and words that should or should not contain biases (equality sets). These lists are based on data collected specifically from the US. Thus, they may not translate to other countries or cultures. Further, some of these vocabulary terms, while peer reviewed, may be subjective and may not fully capture the bias subspace. 

Recent work by \citet{gonen2019lipstick} suggests that debiasing methods based on bias component removal are insufficient to completely remove bias in the embeddings, since embeddings with similar biases are still clustered together after bias component removal. Following \possesivecite{gonen2019lipstick} procedure, we plot the number of neighbors of a particular bias class as a function of the original bias, before and after debiasing in Figure~\ref{fig:bias_plot_religion_jew} and \ref{fig:bias_plot_religion_muslim} in the Appendix. In line with \possesivecite{gonen2019lipstick} findings, simply removing the bias component is insufficient to remove multiclass ``cluster bias''. However, increasing the size of the bias subspace reduces the correlation of the two variables (Table~\ref{tab:correlation_neighbors} in the Appendix).



\section{Conclusion}
We showed that word embeddings trained on \url{www.reddit.com} data contain multiclass biases. We presented a novel metric for evaluating debiasing procedures for word embeddings. We robustly removed multiclass bias using a generalization of existing techniques. Finally, we showed that this multiclass generalization preserves the utility of embeddings for different NLP tasks.

\section*{Acknowledgments}
This research was supported by Grant No.~IIS1812327 from the United States National Science Foundation (NSF). 
We also acknowledge several people who contributed to this work: Benjamin Pall for his valuable early support of this work; Elise Romberger who helped edit this work prior to its final submission. Finally, we are greatly appreciative of the anonymous reviewers for their time and constructive comments. 

\bibliography{naaclhlt2019}
\bibliographystyle{acl_natbib}

\appendix

\section{Addressing Cluster Bias}

To visualize the degree of cluster bias before and after our debiasing procedure, we follow a similar procedure to \citet{gonen2019lipstick}. For a defining set $D$ for the target task (e.g. religion, race, gender), we compute the mean embedding $\bm{\mu} = \frac{1}{|D|} \sum_{\mathbf{c} \in D} \mathbf{c}$. Then, for each class $\mathbf{c}$ in the defining set, we define the bias direction as $\mathbf{b} = \frac{\mathbf{c} - \bm{\mu}}{\lVert \mathbf{c} - \bm{\mu} \rVert}$. Using this, we find the 500 most biased words in each direction in the whole vocabulary based on their component in the bias direction: $\langle \mathbf{w}, \mathbf{b} \rangle$.

\begin{figure}
    \centering
    \begin{subfigure}{\linewidth}
        \includegraphics[width=\linewidth]{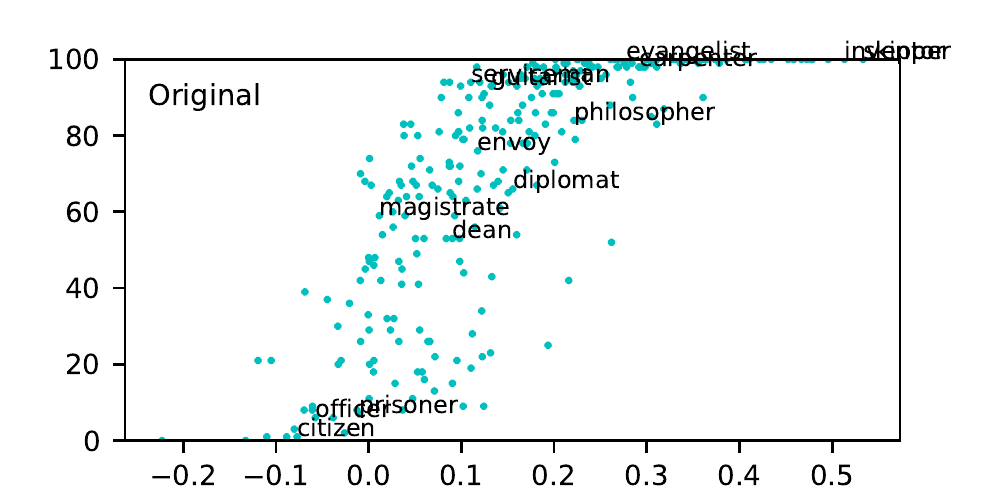}
    \end{subfigure}
    \begin{subfigure}{\linewidth}
        \includegraphics[width=\linewidth]{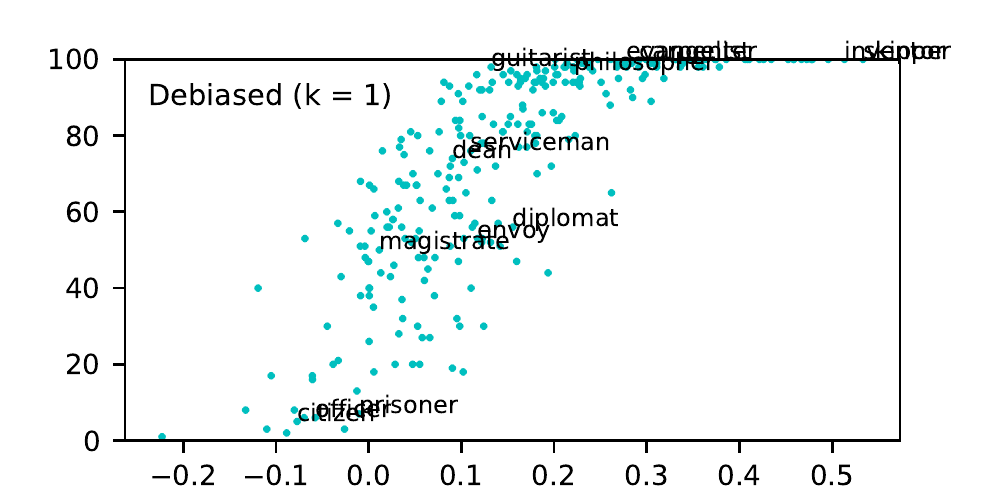}
    \end{subfigure}
    \begin{subfigure}{\linewidth}
        \includegraphics[width=\linewidth]{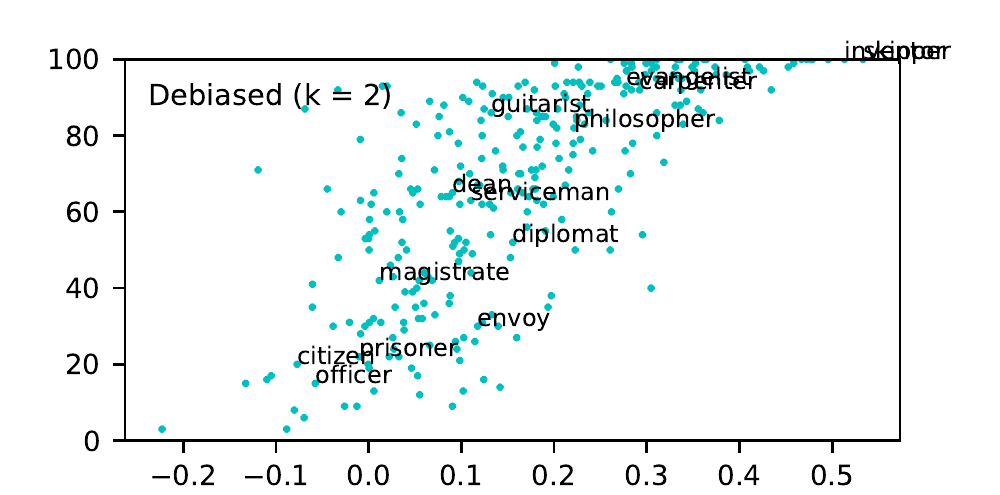}
    \end{subfigure}
    \caption{Plots of number of neighbors to \textit{jew} for each profession as a function of its original bias with respect to \textit{jew}, before and after debiasing, for different subspace dimensionalities $k$.}
    \label{fig:bias_plot_religion_jew}
\end{figure}

\begin{figure}[!t]
    \centering
    \begin{subfigure}{\linewidth}
        \includegraphics[width=\linewidth]{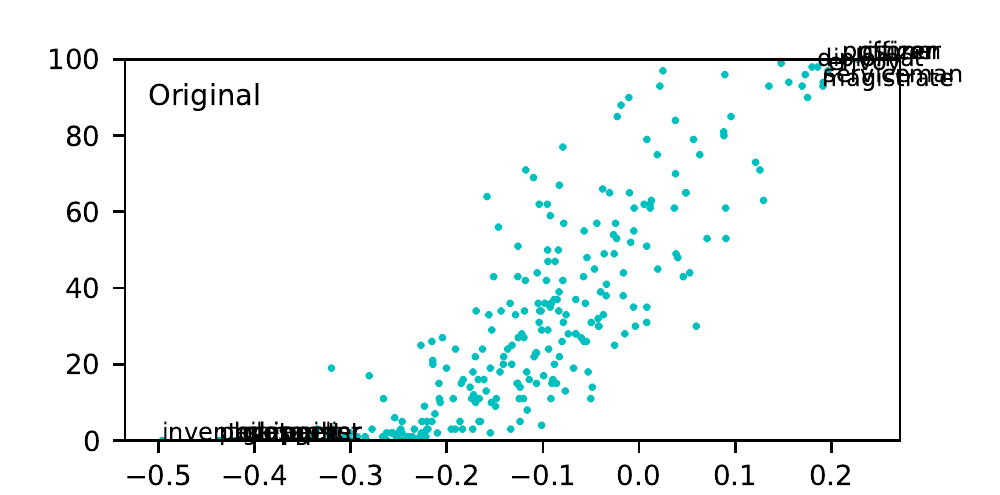}
    \end{subfigure}
    \begin{subfigure}{\linewidth}
        \includegraphics[width=\linewidth]{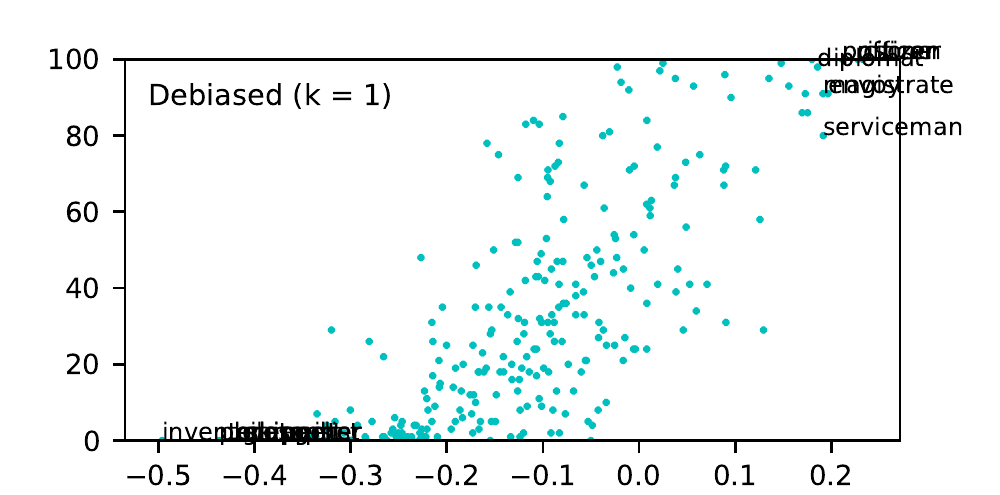}
    \end{subfigure}
    \begin{subfigure}{\linewidth}
        \includegraphics[width=\linewidth]{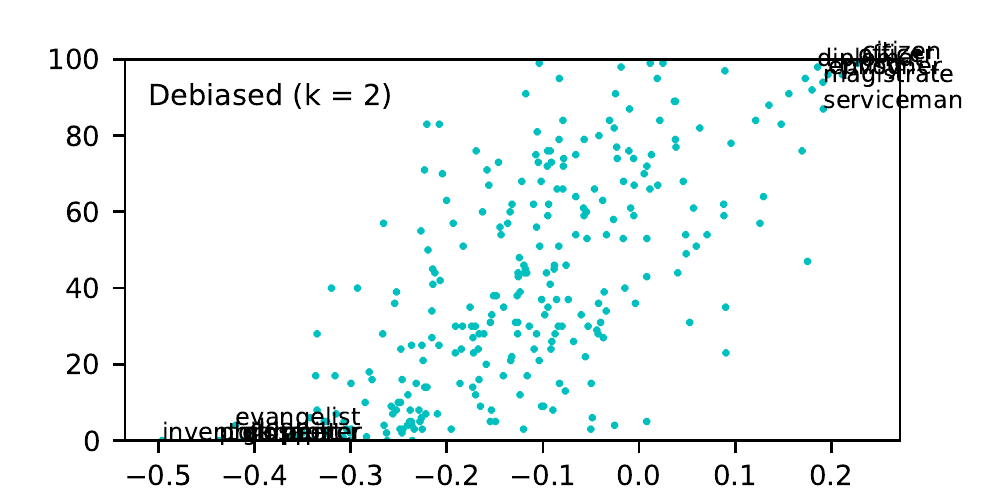}
    \end{subfigure}
    \caption{Plots of number of neighbors to \textit{muslim} for each profession as a function of its original bias with respect to \textit{muslim}, before and after debiasing, for different subspace dimensionalities $k$.}
    \label{fig:bias_plot_religion_muslim}
\end{figure}

Then, using the list of professions from \citet{bolukbasi2016man}\footnote{\url{https://github.com/tolga-b/debiaswe/blob/master/data/professions.json}}, we find the 100 closest neighbors for each profession. We then plot the number of neighbors with positive bias against the original bias of the profession word, as shown in Figures \ref{fig:bias_plot_religion_jew} and \ref{fig:bias_plot_religion_muslim}. The plots suggest that while the correlation between the bias component and the number of positively-biased neighbors might decrease slightly as the number of bias subspace dimensions increase, the cluster bias is still not fully removed. As Table \ref{tab:correlation_neighbors} shows, while the correlation between the two quantities decreases as the number of subspace dimensions increase to 2 or 3, its magnitude is still high.


\begin{table}[]
    \centering
    \begin{tabular}{|c|c|c|c|}
        \hline
        Target & $k$ & $r$ & $\rho$ \\
        \hline
        \multirow{4}{*}{\textit{jew}} & 0 & 0.767 & 0.875 \\
        & 1 & 0.795 & 0.891 \\
        & 2 & 0.718 & 0.756 \\
        & 3 & 0.736 & 0.772 \\
        \hline
        \multirow{4}{*}{\textit{christian}} & 0 & 0.925 & 0.947 \\
        & 1 & 0.835 & 0.841 \\
        & 2 & 0.825 & 0.831 \\
        & 3 & 0.832 & 0.839 \\
        \hline
        \multirow{4}{*}{\textit{muslim}} & 0 & 0.858 & 0.894 \\
        & 1 & 0.774 & 0.812 \\
        & 2 & 0.715 & 0.721 \\
        & 3 & 0.712 & 0.718 \\
        \hline
    \end{tabular}
    \caption{Pearson's $r$ and Spearman's $\rho$ correlation coefficients between the number of biased neighbors and the original bias of professions with respect to target classes for religion. $k$ is the dimension of the bias subspace used ($k = 0$ represents the original embedding). All correlation coefficients have $p$-values $< 10^{-30}$.}
    \label{tab:correlation_neighbors}
\end{table}


\end{document}